\documentclass[conference]{IEEEtran}
\IEEEoverridecommandlockouts
\usepackage{cite}
\usepackage{amsmath,amssymb,amsfonts}
\usepackage{algorithmic}
\usepackage{graphicx}
\usepackage{textcomp}
\usepackage{xcolor}
\usepackage{algorithm}
\usepackage{booktabs}
\usepackage{listings}
\usepackage{hyperref}
\usepackage{multirow}
\usepackage{array}
\usepackage{multirow}
\usepackage{makecell}
\usepackage{tabularx}
\usepackage{graphicx}   
\usepackage{subcaption} 
\renewcommand{\emph}[1]{\textbf{#1}}
\def\BibTeX{{\rm B\kern-.05em{\sc i\kern-.025em b}\kern-.08em
    T\kern-.1667em\lower.7ex\hbox{E}\kern-.125emX}}
\begin{document}

\title{STAR: A Stage-attributed Triage and Repair framework for RCA Agents in Microservices\\
\author{\IEEEauthorblockN{1\textsuperscript{st} Junle Wang}
\IEEEauthorblockA{\textit{School of Artificial Intelligence,} \\
\textit{Beihang University}\\
Beijing, China \\
wjl\_admire@buaa.edu.cn}
\and
\IEEEauthorblockN{2\textsuperscript{nd} Xingchuang Liao}
\IEEEauthorblockA{\textit{School of Artificial Intelligence,} \\
\textit{Beihang University}\\
Beijing, China \\
liaoxingchuang@buaa.edu.cn}
\and
\IEEEauthorblockN{3\textsuperscript{rd} Wenjun Wu}
\IEEEauthorblockA{\textit{School of Artificial Intelligence,} \\
\textit{Beihang University}\\
Beijing, China \\
wwj09315@buaa.edu.cn}
\and
}
\thanks{ }
}

\maketitle

\begin{abstract}
LLM-based root cause analysis (RCA) agents have recently emerged as a promising paradigm for incident diagnosis in microservice AIOps. However, their reliability remains fragile: an error in early evidence collection, hypothesis formulation, or causal analysis can propagate through the reasoning trace and eventually corrupt the final diagnosis. In this paper, we present \textbf{STAR}, a \emph{Stage-attributed Triage and Repair} framework for repairing erroneous RCA traces. STAR explicitly decomposes an RCA workflow into four structured stages, namely \emph{Evidence Package} (EP), \emph{Hypothesis Set} (HS), \emph{Analysis Structure} (AS), and \emph{Decision Report} (DR), and treats agent failure as a stage-localizable reasoning bug rather than a monolithic end-to-end error. Built on top of LangGraph, STAR performs stage-wise auditing, budget-aware \emph{Fast/Slow Routing}, \emph{decisive stage localization via counterfactual candidate evaluation}, and stage-specific patch-and-replay repair.

We evaluate STAR on a public large-scale benchmark and a real-world production dataset, using two RCA agent workflows and three foundation models. Experimental results show that STAR consistently improves both root cause localization and fault type classification over strong baselines. Moreover, STAR identifies the decisive faulty stage with high accuracy, repairs most initially incorrect traces within one or two replay rounds, and benefits substantially from both Fast/Slow Routing and counterfactual stage evaluation. These results suggest that explicitly modeling \emph{where} an RCA agent fails is an effective path toward reliable, debuggable, and self-repairing agentic RCA systems.
\end{abstract}


\section{Introduction}

Microservice architectures have become the dominant paradigm for large-scale cloud applications due to their scalability, flexibility, and support for independent deployment. However, the same decentralization that enables rapid iteration also makes reliability engineering substantially more challenging. A single fault can propagate across services, pods, or nodes, while its observable symptoms often appear far from the true origin. As a result, root cause analysis (RCA) in microservices is both operationally important and intrinsically difficult~\cite{causeinfer2014,facgraph2018,msrank2019,microhecl2021,eadro2023}.

Recent advances in large language models (LLMs) have inspired a new class of \emph{LLM-based RCA agents} that reason over multimodal observability signals---metrics, logs, and traces---to infer root causes and generate diagnostic explanations. Compared with conventional correlation-based or graph-based RCA pipelines, these agents are more adaptable to open-ended environments and more capable of synthesizing heterogeneous evidence~\cite{mabc2024,flowofaction2025,react2023,autogen2023,tot2023}. However, their practical utility remains limited by the fragility of the \emph{reasoning process itself}. In RCA, even a minor error in evidence scoping, hypothesis formulation, or causal interpretation can propagate through later reasoning steps and ultimately lead to an incorrect diagnosis.

This problem is particularly acute in microservice RCA because the task is inherently structured. Correct diagnosis depends not only on textual plausibility, but also on telemetry consistency, causal reachability, temporal ordering, and deployment topology. Consequently, debugging an RCA agent solely from raw free-form reasoning traces is often unreliable and inefficient: such traces are noisy, decisive errors are difficult to isolate, and regenerating long trajectories is costly. More importantly, correcting an isolated reasoning step often fails to address the actual failure source, which typically lies in a higher-level workflow artifact, such as incomplete evidence, biased hypotheses, infeasible causal chains, or unstable final decisions.

These observations motivate a process-centric question: instead of only asking \emph{which service is faulty}, can we also determine \emph{which stage of the RCA workflow is faulty}, and repair the diagnosis by replaying only the affected downstream stages? To answer this question, we propose \textbf{STAR} (\textbf{S}tage-attributed \textbf{T}riage \textbf{A}nd \textbf{R}epair), a debugging and repair layer for LLM-based RCA agents. STAR explicitly decomposes an RCA trace into four structured artifacts: \emph{Evidence Package} (EP), \emph{Hypothesis Set} (HS), \emph{Analysis Structure} (AS), and \emph{Decision Report} (DR). Rather than treating agent failure as a black-box end-to-end error, STAR models it as a \emph{stage-localizable reasoning bug}. It first audits the RCA trace, then identifies the decisive faulty stage, patches the corresponding artifact, and finally replays only the downstream reasoning to eliminate error propagation.

To make this repair process practical, STAR incorporates three key mechanisms. First, it performs \emph{stage-wise audit and diagnosis}, transforming vague failure signals into explicit stage-level inconsistency evidence. Second, it adopts \emph{Fast/Slow Routing} to balance correction cost and effectiveness: lightweight local repair is applied to near-miss traces, while replay-based localization is reserved for strongly contaminated cases. Third, STAR introduces \emph{decisive stage localization via counterfactual candidate evaluation}, in which candidate stage patches are assessed by replaying downstream stages and examining whether the repaired trace improves. This allows STAR to identify not merely a suspicious stage, but the earliest stage whose correction can restore RCA consistency. Built on LangGraph, STAR further leverages node-level replay and structured state artifacts to enable controllable implementation and systematic repair analysis.

Experiments on both a public AIOps benchmark and a real-world production dataset show that STAR consistently improves end-to-end RCA performance across two agent workflows and three foundation models. In particular, STAR substantially improves root cause localization and fault type classification over the original workflows, identifies the decisive faulty stage with high precision, and repairs most initially incorrect traces within one or two replay rounds. Additional ablation studies further show that both Fast/Slow Routing and counterfactual decisive-stage localization contribute significantly to repair efficiency and final diagnostic accuracy.

In summary, this paper makes the following contributions:
\begin{itemize}
    \item We present \textbf{STAR}, a stage-attributed debugging and replay framework for LLM-based RCA agents, which decomposes RCA into four structured stages and supports stage-wise audit, decisive stage localization, patching, and downstream replay.
    \item We design two key mechanisms for effective repair: \emph{Fast/Slow Routing} for budget-aware correction and \emph{counterfactual candidate evaluation} for decisive stage localization.
    \item Extensive experiments on public and real-world datasets show that STAR consistently improves diagnosis quality, stage attribution accuracy, and repair efficiency across datasets, workflows, and backbone models.
\end{itemize}

\section{Preliminaries and Motivation}
\label{sec:prelim}

\subsection{Microservice Root Cause Analysis}
Modern cloud-native applications are increasingly built on microservice architectures, where functionality is decomposed into independently deployable services connected through complex runtime dependencies. While this design improves scalability and agility, it also makes reliability management substantially more difficult: faults can propagate across services, and observed symptoms often appear far from their true origin. As a result, \emph{root cause analysis} (RCA) in microservices aims not only to identify the most likely faulty entity (e.g., host, pod, or service), but also to explain how the failure propagates through the system.

In practice, effective RCA relies on multi-modal observability signals---metrics, logs, and traces---together with dependency information that captures service interactions and deployment structure. Prior studies have shown that robust fault localization in microservices critically depends on jointly reasoning over telemetry and topology, especially under noisy, incomplete, and dynamically changing environments~\cite{nezha2023,mrca2024,tracemd2024,micronet2024,tracenet2023,tracerca2021,chainofevent2024,sparserca2024}. This also makes RCA fundamentally different from generic reasoning tasks: correctness is constrained by evidence, time, and system structure rather than by textual plausibility alone.

\subsection{Motivation}
\subsubsection{Motivation 1}Reasoning Failures in RCA Agents Directly Degrade RCA Accuracy.
These characteristics make LLM-based RCA agents particularly vulnerable to reasoning failures. Errors in evidence scoping (e.g., missing anomaly onset or focusing only on downstream victims), premature hypothesis anchoring, infeasible causal paths, or overconfident final decisions can directly alter the ranked root-cause candidates and therefore degrade RCA accuracy. Unlike general text-generation tasks, where an imperfect intermediate thought may still lead to the correct answer, RCA requires consistent intermediate commitments to telemetry, temporal order, and topology; even a small reasoning defect can therefore propagate into a large end-to-end error.

\subsubsection{Motivation 2}Fine-Grained Step-by-Step Failure Localization Is Unreliable and Inefficient for RCA Agents. A natural response is to debug the agent at the level of individual reasoning steps. However, step-by-step localization is often both unreliable and inefficient in RCA settings. Fine-grained traces are typically noisy, with many redundant or weakly causal steps, making it difficult to determine which step is truly decisive. More importantly, correcting an isolated step rarely fixes the underlying RCA failure, which usually lies in higher-level workflow artifacts such as evidence scope, hypothesis coverage, causal feasibility, or decision calibration. Repeatedly inspecting and regenerating long reasoning traces also introduces substantial LLM and tool-call overhead. These limitations motivate a more practical alternative: \emph{stage-level} localization and replay-based repair, which targets the decisive faulty stage in the RCA workflow and corrects downstream reasoning through structured patching and replay.


\section{Problem Statement}
\label{sec:problem}

Following Sec.~\ref{sec:prelim}, we formulate microservice RCA as a multi-stage reasoning process over multi-modal observability and system topology. Instead of treating the RCA agent as a black-box predictor, we represent its execution as a structured \emph{program trace}:
\begin{equation}
\mathcal{A} = (\mathrm{EP}, \mathrm{HS}, \mathrm{AS}, \mathrm{DR}),
\end{equation}
where each stage corresponds to a distinct intermediate artifact in the RCA workflow.

\paragraph{Evidence Package (EP)}
$\mathrm{EP}$ defines the incident time window, the entity scope under analysis (including host/service/pod mappings), and an indexed set of evidence items extracted from observability signals. Each evidence item is associated with an identifier, modality, target entity or edge, and a compact summary.

\paragraph{Hypothesis Set (HS)}
$\mathrm{HS}$ consists of a set of candidate explanations $\{h_i\}$ for the incident. Each hypothesis is explicitly grounded in the relevant entities and supporting evidence identifiers from EP, preventing unsupported or arbitrarily chosen evidence from implicitly driving the diagnosis.

\paragraph{Analysis Structure (AS)}
$\mathrm{AS}$ captures the agent's causal reasoning as a set of propagation paths $\{p_j\}$. Each path is represented as a topology-consistent walk or subgraph over the system graph, together with textual justification and supporting evidence identifiers. This formulation makes causal reasoning verifiable in terms of reachability, temporal consistency, and evidential support.

\paragraph{Decision Report (DR)}
$\mathrm{DR}$ outputs a ranked list of root-cause candidates with confidence scores, along with minimal verification tests or recommended actions. When uncertainty remains high, DR may favor a verification-first conclusion over an overconfident localization.

We formulate agent self-repair as \emph{stage-attributed correction with replay}. Let
\begin{equation}
s \in \{S_1, S_2, S_3, S_4\}
\end{equation}
denote the stage index corresponding to EP, HS, AS, and DR, respectively. A stage patch operator generates a corrected artifact at stage $s$:
\begin{equation}
\mathcal{O}'(s) = \mathcal{P}_s(\mathcal{A}, O, G),
\end{equation}
where $O$ denotes the incident observability and $G$ denotes the system topology. Given a patched artifact, we define a deterministic replay operator that re-executes all downstream stages:
\begin{equation}
\mathrm{Replay}(\mathcal{A}, s),
\end{equation}
which replaces the stage-$s$ artifact with its patched version and reruns all subsequent stages through $S_4$.

The central challenge is to identify the \emph{decisive faulty stage}. We define $s^*$ as the earliest stage such that patching its artifact and replaying all downstream stages yields a substantial improvement in trajectory reliability and/or final RCA correctness, while patching only later stages cannot consistently achieve the same effect. This definition captures the stage-wise contamination property of RCA agents: once an upstream artifact is flawed, downstream hypotheses, analyses, and decisions may remain systematically biased unless replay is initiated from the corrected upstream stage.

Accordingly, given an initial RCA trace $\mathcal{A}$, our objective is to:
\begin{enumerate}
    \item determine whether the trace is unreliable,
    \item identify the decisive faulty stage $s^*$,
    \item patch only the artifact at stage $s^*$, and
    \item replay the downstream stages to obtain a repaired decision report.
\end{enumerate}

\section{Methodology}
\label{sec:method}
\begin{figure}[htbp]
    \centering
    \includegraphics[width=\linewidth]{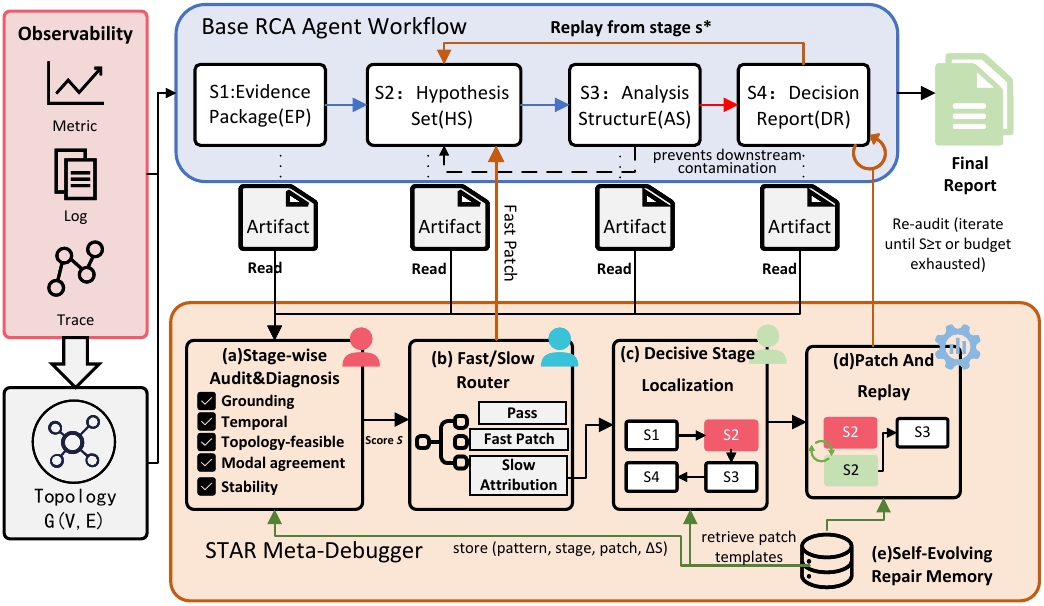}
    \caption{Overview of our proposed framework \textbf{STAR}.}
    \label{fig:overall}
\end{figure}
STAR consists of five tightly coupled components: \emph{stage-wise audit}, \emph{fast/slow routing}, \emph{decisive stage localization}, \emph{patch-and-replay repair} and \emph{self-Evolving Repair Memory}. As shown in Fig.~\ref{fig:overall}, we propose \textbf{STAR} as a process-centric reliability layer for microservice RCA agents. Building on the stage-structured formulation in Sec.~\ref{sec:problem}, STAR does not re-solve RCA from scratch; instead, it audits the RCA trace, identifies the decisive faulty stage, repairs the corresponding stage artifact, and replays only the downstream stages to remove error contamination. This design is guided by three RCA-specific requirements: intermediate reasoning must remain grounded in observability signals, causal explanations must be consistent with service topology and temporal order, and the final diagnosis must remain operationally actionable.

\subsection{Stage-wise Audit and Diagnosis}
\label{subsec:method_audit}

Given an RCA trace $\mathcal{A}=(\mathrm{EP},\mathrm{HS},\mathrm{AS},\mathrm{DR})$, STAR first performs an RCA-oriented audit to determine both \emph{whether} the trace is unreliable and \emph{where} the inconsistency first emerges. The audit outputs a global reliability score $S$ together with a set of stage diagnostics
\begin{equation}
\mathbf{d}=\{d_{S1}, d_{S2}, d_{S3}, d_{S4}\},
\end{equation}
where each $d_{Si}$ records violated constraints, severity, and blame evidence for stage $S_i$.

Rather than assigning confidence in a generic way, the audit is tied to the structural commitments of RCA. For the evidence package, STAR checks whether the selected time window covers anomaly onset, whether modality coverage matches the incident type, and whether the entity scope includes plausible upstream/downstream neighborhoods rather than only downstream victims. For the hypothesis set, STAR verifies that each hypothesis is grounded in EP evidence, that the search space is not prematurely collapsed into a single anchored explanation, and that host/service/pod interactions are considered when implied by the evidence. For the analysis structure, STAR checks whether each causal path is reachable in $G$, whether anomaly onsets respect cause-before-effect, and whether intermediate links are supported by telemetry. For the decision report, STAR examines whether confidence is calibrated to evidence sufficiency, whether the ranking is consistent with the preceding analysis, and whether recommended tests are discriminative and mechanism-consistent.

These local checks are aggregated into a global score
\begin{equation}
S=\sum_k w_k \, s_k,
\qquad \sum_k w_k=1,
\end{equation}
Here, $s_k \in [0,1]$ denotes the normalized score of the $k$-th audit criterion, and $w_k$ is its importance weight. A higher $S$ indicates that the RCA trace is more self-consistent and better satisfies the RCA-specific requirements. In this way, STAR transforms a vague signal that ``the agent is wrong'' into a concrete diagnosis of \emph{which stage violates RCA requirements and why}.

\subsection{Fast/Slow Routing for Budget-Aware Repair}
\label{subsec:method_routing}

Not all faulty traces require the same level of intervention. In practice, some are \emph{near-miss} cases, where the overall trajectory remains largely valid and only a local inconsistency appears in a downstream artifact, while others reflect \emph{systemic breakdowns}, where upstream errors in evidence scoping or hypothesis construction have already contaminated the entire downstream reasoning chain. Applying the same repair procedure to both cases would be inefficient: full replay-based debugging is unnecessarily expensive for near-miss traces, while local patching is ineffective once upstream contamination has occurred.

To balance repair effectiveness and cost, STAR introduces a Fast/Slow routing mechanism based on the audit score:

\begin{equation}
\mathrm{Routing}=
\begin{cases}
\text{Pass}, & \text{if } S \ge \tau,\\
\text{Fast Path}, & \text{if } \tau-\epsilon \le S < \tau,\\
\text{Slow Path}, & \text{if } S < \tau-\epsilon.
\end{cases}
\end{equation}
If $S \ge \tau$, the trace is considered sufficiently reliable and the original decision is accepted. If $S$ falls into the intermediate region $[\tau-\epsilon,\tau)$, STAR activates the \textbf{Fast Path}. This regime is designed for traces that are mostly correct but contain a local defect. STAR directly selects the stage with the highest diagnostic severity, applies a single constrained patch, and replays only the necessary downstream stages. Fast Path therefore serves as a lightweight repair mode for small but operationally important inconsistencies, such as confidence miscalibration in DR or minor infeasibility in AS.

When $S < \tau-\epsilon$, STAR enters the \textbf{Slow Path}, which is reserved for traces with strong evidence of upstream contamination. Such traces typically exhibit hard RCA violations, including missing anomaly onset in EP, insufficient modality coverage, severe hypothesis anchoring, or widespread topology/temporal conflicts in AS. In this setting, local patching is unlikely to be sufficient, because the downstream artifacts have already been generated from a corrupted upstream state. STAR therefore switches to replay-validated stage localization.

\subsection{Decisive Stage Localization via Counterfactual Candidate Evaluation}
\label{subsec:method_attrib}

The central challenge in STAR is to identify the \emph{decisive faulty stage}, i.e., the earliest stage whose correction can repair the trace after downstream replay. This design directly follows the stage-dependency contamination discussed in Sec.~\ref{sec:problem}: if an upstream artifact is already flawed, later stages may remain coherently wrong and cannot be fixed reliably in isolation.

\begin{figure}[htbp]
    \centering
    \includegraphics[width=\linewidth]{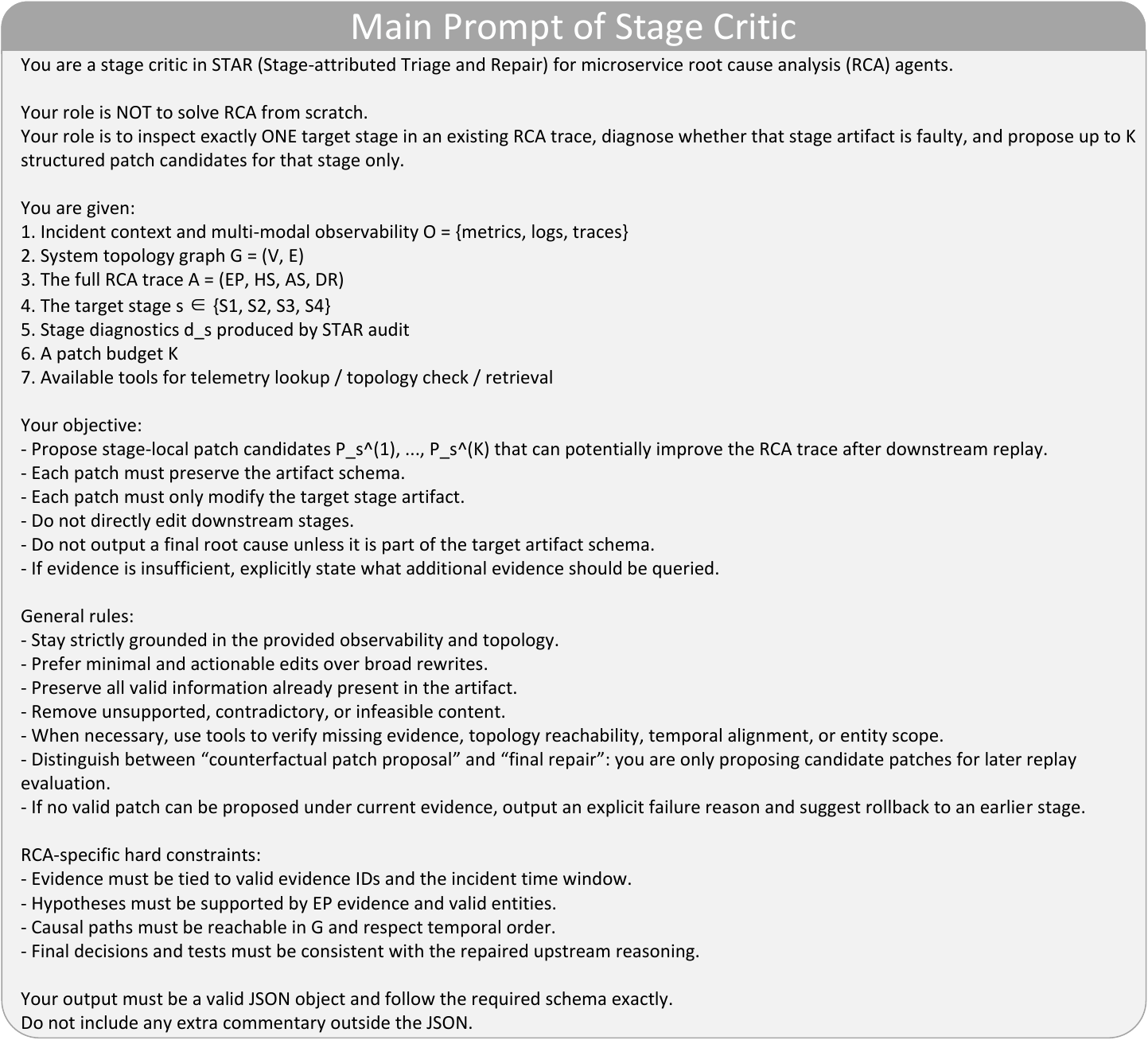}
    \caption{Main prompt of stage critic $\mathcal{C}_s$.}
    \label{fig:prompt}
\end{figure}

For each stage $s$, STAR invokes a stage critic $\mathcal{C}_s$(main prompt is shown in Fig.~\ref{fig:prompt}) to propose $K$ structured patch candidates:
\begin{equation}
\{\mathcal{P}_s^{(k)}\}_{k=1}^{K} = \mathcal{C}_s(\mathcal{A}, d_s; O,G).
\end{equation}
Here, patching is only used as a counterfactual probe for stage attribution; the selected patch is not committed until the decisive stage is identified. Each candidate is then evaluated by applying patch and replay, followed by re-audit:
\begin{equation}
\Delta S(s,k) = S\!\left(\mathrm{Replay}(\mathcal{A}, s, \mathcal{P}_s^{(k)})\right) - S(\mathcal{A}).
\end{equation}
STAR searches stages in causal order and selects the earliest stage that yields a significant improvement:
\begin{equation}
s^* = \min \left\{ s \;\middle|\; \max_k \Delta S(s,k) \ge \delta \right\},
\end{equation}
where $\delta$ is a minimum improvement margin. This replay-validated criterion makes stage attribution both testable and actionable: STAR does not merely identify a suspicious stage, but the earliest stage whose correction can actually restore trace consistency.

To avoid wasting repair budget on downstream reasoning when the evidence itself is insufficient, STAR further introduces an exception-driven rollback rule. If the analysis stage cannot produce any topology-feasible and telemetry-supported causal chain under hard constraints, STAR triggers an \textsc{InsufficientEvidenceException} and forces rollback to S1, thereby re-entering evidence recollection rather than overfitting a flawed downstream explanation.

\subsection{RCA-Specific Patch-and-Replay Repair}
\label{subsec:method_repair}
\begin{table*}[t]
\centering
\caption{Summary of STAR patch operators across different RCA stages. Each failure pattern is paired with its corresponding repair action, followed by replay of downstream stages to remove contamination.}
\label{tab:star_patch_summary}
\footnotesize
\setlength{\tabcolsep}{3.5pt}
\renewcommand{\arraystretch}{1.15}

\begin{tabularx}{0.85\textwidth}{
>{\centering\arraybackslash}m{0.06\textwidth}
>{\centering\arraybackslash}m{0.14\textwidth}
>{\raggedright\arraybackslash}m{0.20\textwidth}
>{\raggedright\arraybackslash}m{0.25\textwidth}
>{\centering\arraybackslash}m{0.11\textwidth}}
\toprule
\textbf{Stage} & \textbf{Artifact} & \textbf{Typical failure pattern} & \textbf{Patch operation} & \textbf{Replay scope} \\
\midrule

\multirow{4}{*}{\textbf{S1}}
& \multirow{4}{=}{\centering Evidence Package (EP)}
& \makecell[l]{Missed anomaly onset}
& \makecell[l]{Shift or expand the incident window}
& \multirow{4}{=}{\centering Replay\\S2--S4} \\
& & \makecell[l]{Missing modality evidence}
& \makecell[l]{Re-query the missing modality}
& \\
& & \makecell[l]{Victim-only scope}
& \makecell[l]{Expand to upstream/downstream neighbors}
& \\
& & \makecell[l]{Misaligned time windows}
& \makecell[l]{Realign telemetry timestamps}
& \\
\midrule

\multirow{4}{*}{\textbf{S2}}
& \multirow{4}{=}{\centering Hypothesis Set (HS)}
& \makecell[l]{Unsupported hypotheses}
& \makecell[l]{Remove unsupported hypotheses}
& \multirow{4}{=}{\centering Replay\\S3--S4} \\
& & \makecell[l]{Anchoring bias}
& \makecell[l]{Introduce alternative candidates}
& \\
& & \makecell[l]{Missing counter-hypotheses}
& \makecell[l]{Add counter-hypotheses}
& \\
& & \makecell[l]{Lack of cross-layer candidates}
& \makecell[l]{Add host--pod--service candidates}
& \\
\midrule

\multirow{4}{*}{\textbf{S3}}
& \multirow{4}{=}{\centering Analysis Structure (AS)}
& \makecell[l]{Unreachable causal paths}
& \makecell[l]{Rebuild a reachable causal chain}
& \multirow{4}{=}{\centering Replay\\S4} \\
& & \makecell[l]{Hallucinated edges}
& \makecell[l]{Prune unsupported edges}
& \\
& & \makecell[l]{Temporal order violations}
& \makecell[l]{Restore cause-before-effect ordering}
& \\
& & \makecell[l]{Unsupported causal links}
& \makecell[l]{Add telemetry support for each link}
& \\
\midrule

\multirow{4}{*}{\textbf{S4}}
& \multirow{4}{=}{\centering Decision Report (DR)}
& \makecell[l]{Overconfident ranking}
& \makecell[l]{Lower or recalibrate confidence}
& \multirow{4}{=}{\centering No replay} \\
& & \makecell[l]{Inconsistent top candidate}
& \makecell[l]{Align ranking with repaired analysis}
& \\
& & \makecell[l]{Weak verification tests}
& \makecell[l]{Replace with discriminative tests}
& \\
& & \makecell[l]{Mechanism-irrelevant actions}
& \makecell[l]{Match actions to the failure mechanism}
& \\
\bottomrule
\end{tabularx}
\end{table*}

Once the decisive stage $s^*$ is identified, STAR applies a stage-local patch and replays downstream stages to obtain a repaired trace $\mathcal{A}^*$ and decision $\mathrm{DR}^*$. Patch operators(shown in Table.~\ref{tab:star_patch_summary}) are constrained to preserve artifact schemas and are designed around RCA-specific primitives: telemetry queries, entity scopes, topology neighborhoods, and verification actions.

For \textbf{EP}, STAR performs evidence recollection under RCA heuristics, including time-window adjustment to recover anomaly onset, expansion to upstream/downstream topology neighbors, rebalancing modality coverage, and re-alignment across telemetry sources. For \textbf{HS}, STAR repairs hypothesis coverage by introducing counter-hypotheses under ambiguous evidence, enforcing host--pod--service consistency, and requiring explicit evidence binding to prevent unsupported speculation. For \textbf{AS}, STAR reconstructs topology-feasible, telemetry-closed causal chains by pruning unreachable edges, restoring cause-before-effect order, and requiring each causal link to be supported by at least one evidence item. For \textbf{DR}, STAR calibrates confidence and improves actionability, preferring verification-first outputs when ambiguity remains high and ensuring that recommended tests match the hypothesized failure mechanism.

\textbf{Replay} is the key mechanism for removing RCA contamination. After patching stage $s^*$, STAR re-executes all downstream stages conditioned on the corrected upstream artifact:
\begin{equation}
\mathrm{Replay}(\mathcal{A}, s, \mathcal{P}_s)
= \mathrm{Run}_{(s+1)\rightarrow 4}\Bigl(\mathrm{Replace}(\mathcal{A}, s, \mathcal{P}_s)\Bigr),
\end{equation}
where $\mathrm{Run}_{(s+1)\rightarrow 4}$ denotes executing the base agent from stage $s+1$ to $S4$ with fixed upstream artifacts. Concretely, patching S1 triggers replay of S2--S4, patching S2 triggers replay of S3--S4, patching S3 triggers replay of S4, and patching S4 requires no replay. This ensures that downstream artifacts are regenerated from the corrected upstream state rather than being locally edited on top of a contaminated trace.

\subsection{Self-Evolving Repair Memory}
\label{subsec:method_memory_loop}

To reduce repeated trial-and-error, STAR maintains a repair memory
\begin{equation}
\mathcal{M}=\left\{\langle \mathbf{q}_i, s_i, \mathcal{P}_i, \Delta S_i \rangle \right\},
\end{equation}
where $\mathbf{q}_i$ summarizes the incident pattern in RCA terms (e.g., dominant anomaly type, modality availability, topology neighborhood statistics, and evidence sparsity), $s_i$ is the blamed stage, $\mathcal{P}_i$ is the patch template, and $\Delta S_i$ is the achieved improvement. This memory serves two purposes: it provides an attribution prior for recurring RCA failure patterns, and it seeds the stage critics with historically successful patch templates to improve repair efficiency.

Overall, STAR iterates for at most $I$ rounds. It first runs the base agent to obtain $\mathcal{A}$, audits the trace to compute $(S,\mathbf{d})$, and either accepts the decision, performs a lightweight Fast-Path patch, or enters Slow-Path replay-validated stage localization. After patching the blamed stage and replaying downstream stages, STAR updates memory if the repair yields sufficient improvement. If the repair budget is exhausted, STAR returns a conservative verification-first output, such as top-$K$ candidates with discriminative tests, to avoid overconfident misdiagnosis.

\section{Experiments}
\label{sec:experiments}
\subsection{Research Questions}
\begin{itemize}
\item RQ1: Does STAR improve end-to-end RCA performance compared with the base RCA agent baselines?

\item RQ2: Can STAR identify the decisive faulty stage (S1–S4) precisely?

\item RQ3: Given an initially incorrect trace, how many repair iterations/replays does STAR typically require to correct the diagnosis?

\item RQ4: How much does STAR benefit from its key components, particularly 
\emph{Fast/Slow Routing} and \emph{Decisive Stage Localization via Counterfactual Candidate Evaluation}?

\end{itemize}

\subsection{Experiment Setup}
\subsubsection{Dataset and Preprocessing}
Dataset A is a large-scale public benchmark released by the AIOps Challenge~\cite{aiopsbench2024}. It was constructed through controlled fault injection on a realistically deployed microservice system, HipsterShop2. The benchmark provides multimodal observability data, including metrics, logs, and traces. HipsterShop2 runs on a dynamic Kubernetes (K8s) cluster consisting of 10 services with 4 replicas per service, yielding 40 pods in total, which are dynamically scheduled across 6 nodes. In total, Dataset A contains 15 fault types, including 9 service-/pod-level faults in the K8s container environment and 6 node-level faults, such as abrupt memory pressure, disk space exhaustion, disk I/O anomalies, CPU contention, and progressive CPU slowdown.

Dataset B is a real-world production dataset collected from a Project Management Platform operated by an electric power information enterprise. In contrast to Dataset A, the incidents in Dataset B were recorded under natural production conditions rather than generated through fault injection. The platform consists of 12 microservices and 48 pods, and the dataset captures multimodal observability signals, including metrics, logs, and traces, during real incident scenarios. The observed faults fall into five categories: CPU hog, memory leak, network delay, packet loss, and disk payload overload.

\subsubsection{Baselines}
To evaluate the generalization ability of \textbf{STAR}, we adapt it to two representative RCA agent systems, \textbf{mABC} and \textbf{RCAgent}.

\textbf{mABC} is a multi-agent RCA framework for microservice systems that organizes diagnosis through a predefined workflow and a set of specialized agents, including alert receiving, process scheduling, data collection, dependency exploration, probability estimation, fault mapping, and solution generation. To improve the reliability of LLM-based diagnosis, mABC further introduces a blockchain-inspired voting mechanism, where multiple agents collaboratively verify intermediate answers and revise low-quality outputs through weighted consensus~\cite{mabc2024}. This design makes mABC a representative multi-agent baseline for structured RCA with explicit coordination and reflection.

\textbf{RCAgent} represents a lighter agentic RCA workflow that performs root cause reasoning by directly integrating observability evidence with LLM-based stepwise diagnosis, without the stronger multi-agent coordination and voting process used in mABC. Compared with mABC, RCAgent provides a weaker but more streamlined reasoning baseline, which makes it suitable for testing whether STAR can also improve simpler agent pipelines. In our implementation, both workflows are re-expressed under the same LangGraph execution interface for fair comparison.

\subsubsection{Evaluation Metrics}
We evaluate the model from two aspects: \emph{root cause localization} and \emph{fault type classification}.

\paragraph{Root Cause Localization.}
We report \textbf{Acc@1}, \textbf{Acc@3}, and \textbf{Acc@5}, which measure whether the ground-truth root cause appears in the top-1, top-3, and top-5 predicted candidates, respectively:
\begin{equation}
\mathrm{Acc@}k=\frac{1}{N}\sum_{i=1}^{N}\mathbf{1}\!\left(r_i\in \hat{\mathcal{R}}_i^{(k)}\right), \quad k\in\{1,3,5\},
\end{equation}
where $N$ is the number of test incidents, $r_i$ is the ground-truth root cause, and $\hat{\mathcal{R}}_i^{(k)}$ denotes the top-$k$ predicted candidates.

\paragraph{Fault Type Classification.}
For fault type prediction, we report micro- and macro-averaged precision, recall, and F1-score, including \textbf{MiPr}, \textbf{MaPr}, \textbf{MiRe}, \textbf{MaRe}, \textbf{MiF1}, and \textbf{MaF1}. For each class $c$, let $TP_c$, $FP_c$, and $FN_c$ denote the numbers of true positives, false positives, and false negatives, respectively. Then
\begin{equation}
\begin{aligned}
P_c&=\frac{TP_c}{TP_c+FP_c},\qquad
R_c=\frac{TP_c}{TP_c+FN_c},\\
F1_c&=\frac{2P_cR_c}{P_c+R_c}.
\end{aligned}
\end{equation}
The macro-averaged metrics are defined as
\begin{equation}
\begin{aligned}
\mathrm{MaPr}&=\frac{1}{C}\sum_{c=1}^{C}P_c,\qquad
\mathrm{MaRe}=\frac{1}{C}\sum_{c=1}^{C}R_c,\\
\mathrm{MaF1}&=\frac{1}{C}\sum_{c=1}^{C}F1_c.
\end{aligned}
\end{equation}
where $C$ is the number of fault categories.
The micro-averaged metrics are computed by aggregating statistics over all classes:
\begin{equation}
\begin{aligned}
\mathrm{MiPr}&=\frac{\sum_c TP_c}{\sum_c(TP_c+FP_c)},\qquad
\mathrm{MiRe}=\frac{\sum_c TP_c}{\sum_c(TP_c+FN_c)},\\
\mathrm{MiF1}&=\frac{2\cdot \mathrm{MiPr}\cdot \mathrm{MiRe}}
{\mathrm{MiPr}+\mathrm{MiRe}}.
\end{aligned}
\end{equation}

By analyzing the agent reasoning traces throughout the RCA process, we identify 13 distinct reasoning failure types spanning four stages(shown in Table~\ref{tab:stage_fault_taxonomy}). To evaluate the accuracy of STAR’s decisive stage localization, we adopt an \emph{LLM-as-a-Judge} protocol~\cite{geval2023,mtbench2023,notfaireval2024}.
Specifically, \textbf{GPT-5.2} serves as an independent evaluator for verifying whether STAR correctly identifies the faulty stage in an RCA trace.
Given the RCA reasoning trajectory and STAR’s predicted stage-level audit result, the judge determines whether the predicted faulty stage is consistent with the trace evidence and with the semantics of the four-stage RCA taxonomy (\textbf{EP/HS/AS/DR}).
To reduce self-enhancement bias, the judge model is kept different from the backbone models used in the evaluated RCA agents.
The evaluation prompt is carefully structured to include:
(1) the definitions, failure criteria, and representative examples of the four RCA stages;
(2) a decomposed reasoning procedure that asks the judge to first inspect the trace and then localize the inconsistency to one of the four stages; and
(3) a fixed output format requiring a stage label, a correctness judgment on STAR’s audit, and a brief evidence-grounded rationale.

\subsubsection{Implementation Details.}
We re-implement all baseline RCA agents under the \textbf{LangGraph} framework to support unified state management and controllable replay.
The main motivation is that LangGraph provides a native node-level execution and replay mechanism, which is essential for our stage-attributed correction setting.
By representing the RCA workflow as a structured execution graph, we can explicitly track intermediate stage artifacts, replay selected nodes, and analyze how local stage repairs affect downstream reasoning outcomes.

To ensure a fair comparison and eliminate the influence of a specific backbone model, each workflow is instantiated with three foundation models: \textbf{GPT-5}, \textbf{Qwen3-Max}, and \textbf{Gemini-2.5-Pro}.
In other words, all workflows are evaluated under the same set of LLM backbones, and the reported results reflect the combined effects of workflow design and stage-level correction rather than the advantage of any single model.
Unless otherwise specified, all other implementation settings are kept identical across workflows. All experiments are conducted on a
server equipped with an NVIDIA A100 80GB GPU and
256GB RAM.

\begin{table*}[t]
\centering
\caption{Main results on root cause localization and fault type classification across different datasets, workflows, and foundation models.}
\label{tab:main_results_star_models}
\scriptsize
\setlength{\tabcolsep}{4pt}
\renewcommand{\arraystretch}{1.12}
\resizebox{\textwidth}{!}{
\begin{tabular}{c c c|ccc|cccccc}
\toprule
\multirow{2}{*}{Dataset} & \multirow{2}{*}{Workflow} & \multirow{2}{*}{Model}
& \multicolumn{3}{c|}{Root Cause Localization}
& \multicolumn{6}{c}{Fault Types Classification} \\
\cmidrule(lr){4-6}\cmidrule(lr){7-12}
& & & Acc@1 & Acc@3 & Acc@5 & MiPr & MaPr & MiRe & MaRe & MiF1 & MaF1 \\
\midrule

\multirow{12}{*}{$A$}
& \multirow{3}{*}{mABC}
& GPT-5         & 37.20\% & 45.10\% & 47.80\% & 0.4870 & 0.4580 & 0.5360 & 0.4970 & 0.5103 & 0.4767 \\
& & Gemini-2.5pro & 35.40\% & 43.30\% & 46.00\% & 0.4720 & 0.4460 & 0.5180 & 0.4870 & 0.4939 & 0.4656 \\
& & Qwen3-max    & 36.20\% & 44.10\% & 46.80\% & 0.4840 & 0.4540 & 0.5240 & 0.4870 & 0.5032 & 0.4699 \\
\cmidrule(lr){2-12}
& \multirow{3}{*}{mABC+STAR}
& GPT-5         & \textbf{56.20\%} & \textbf{66.10\%} & \textbf{70.80\%} & \textbf{0.6940} & \textbf{0.6640} & \textbf{0.7490} & \underline{0.7110} & \textbf{0.7204} & \underline{0.6867} \\
& & Gemini-2.5pro & 53.70\% & 63.80\% & 68.40\% & 0.6760 & 0.6510 & 0.7310 & 0.6990 & 0.7024 & 0.6741 \\
& & Qwen3-max    & \underline{54.90\%} & \underline{65.00\%} & \underline{69.60\%} & \underline{0.6890} & \underline{0.6630} & \underline{0.7420} & \textbf{0.7210} & \underline{0.7145} & \textbf{0.6908} \\
\cmidrule(lr){2-12}
& \multirow{3}{*}{RCAgent}
& GPT-5         & 23.80\% & 30.10\% & 32.60\% & 0.3470 & 0.3200 & 0.4090 & 0.3640 & 0.3755 & 0.3406 \\
& & Gemini-2.5pro & 22.00\% & 28.30\% & 30.80\% & 0.3320 & 0.3080 & 0.4010 & 0.3540 & 0.3633 & 0.3294 \\
& & Qwen3-max    & 22.80\% & 29.10\% & 31.60\% & 0.3440 & 0.3160 & 0.3970 & 0.3540 & 0.3686 & 0.3339 \\
\cmidrule(lr){2-12}
& \multirow{3}{*}{RCAgent+STAR}
& GPT-5         & 48.70\% & 58.40\% & 62.50\% & 0.6260 & 0.5930 & 0.6880 & 0.6420 & 0.6555 & 0.6165 \\
& & Gemini-2.5pro & 46.10\% & 55.80\% & 59.90\% & 0.6080 & 0.5770 & 0.6670 & 0.6260 & 0.6361 & 0.6008 \\
& & Qwen3-max    & 47.40\% & 57.10\% & 61.20\% & 0.6210 & 0.5890 & 0.6810 & 0.6450 & 0.6496 & 0.6157 \\
\midrule

\multirow{12}{*}{$B$}
& \multirow{3}{*}{mABC}
& GPT-5         & 38.70\% & 47.00\% & 51.10\% & 0.5420 & 0.5150 & 0.5890 & 0.5540 & 0.5645 & 0.5338 \\
& & Gemini-2.5pro & 36.90\% & 45.20\% & 49.30\% & 0.5270 & 0.5030 & 0.5710 & 0.5440 & 0.5481 & 0.5227 \\
& & Qwen3-max    & 37.70\% & 46.00\% & 50.10\% & 0.5390 & 0.5110 & 0.5770 & 0.5440 & 0.5574 & 0.5270 \\
\cmidrule(lr){2-12}
& \multirow{3}{*}{mABC+STAR}
& GPT-5         & \textbf{58.20\%} & \textbf{68.60\%} & \textbf{73.90\%} & \textbf{0.7420} & \textbf{0.7120} & \textbf{0.7910} & \underline{0.7560} & \textbf{0.7657} & \underline{0.7333} \\
& & Gemini-2.5pro & 55.60\% & 65.90\% & 71.20\% & 0.7240 & 0.6950 & 0.7720 & 0.7390 & 0.7472 & 0.7163 \\
& & Qwen3-max    & \underline{57.00\%} & \underline{67.30\%} & \underline{72.60\%} & \underline{0.7370} & \underline{0.7060} & \underline{0.7840} & \textbf{0.7640} & \underline{0.7598} & \textbf{0.7339} \\
\cmidrule(lr){2-12}
& \multirow{3}{*}{RCAgent}
& GPT-5         & 26.60\% & 33.40\% & 36.90\% & 0.4060 & 0.3780 & 0.4590 & 0.4260 & 0.4309 & 0.4006 \\
& & Gemini-2.5pro & 24.80\% & 31.60\% & 35.10\% & 0.3910 & 0.3660 & 0.4410 & 0.4160 & 0.4145 & 0.3894 \\
& & Qwen3-max    & 25.60\% & 32.40\% & 35.90\% & 0.4030 & 0.3740 & 0.4470 & 0.4160 & 0.4239 & 0.3939 \\
\cmidrule(lr){2-12}
& \multirow{3}{*}{RCAgent+STAR}
& GPT-5         & 50.10\% & 60.70\% & 65.20\% & 0.6740 & 0.6450 & 0.7230 & 0.6890 & 0.6976 & 0.6663 \\
& & Gemini-2.5pro & 47.50\% & 57.90\% & 62.30\% & 0.6550 & 0.6280 & 0.7010 & 0.6720 & 0.6772 & 0.6493 \\
& & Qwen3-max    & 48.90\% & 59.40\% & 63.90\% & 0.6680 & 0.6410 & 0.7150 & 0.6910 & 0.6907 & 0.6652 \\
\bottomrule
\end{tabular}
}
\end{table*}

\begin{table*}[t]
\centering
\caption{Stage-level reasoning fault taxonomy used in STAR auditing and LLM-as-a-Judge evaluation.}\label{tab:stage_fault_taxonomy}
\small
\setlength{\tabcolsep}{5pt}
\renewcommand{\arraystretch}{1.08}
\begin{tabular}{llc}
\toprule
\textbf{Name} & \textbf{Description} & \textbf{Stage} \\
\midrule

\textbf{Fabricated evidence}
& The agent cites non-existent alerts, metrics, logs, traces, or tool outputs.
& \multirow{4}{*}{\centering\parbox[c]{0.14\textwidth}{\centering Evidence Package}} \\

\textbf{Evidence misreading}
& The agent misinterprets evidence semantics, such as metric trends, log meaning.
& \\

\textbf{Source confusion}
& The agent confuses the symptom-observing component with the true fault source.
& \\

\textbf{Biased evidence selection}
& The agent overlooks more diagnostic clues and selects evidence unsystematically.
& \\

\midrule

\textbf{Premature anchoring}
& The agent fixates too early on one candidate and ignores alternatives.
& \multirow{3}{*}{\centering\parbox[c]{0.14\textwidth}{\centering Hypothesis set}} \\

\textbf{Over-specific hypothesis}
& The agent proposes an overly specific fault hypothesis without enough evidence.
& \\

\textbf{Missing hypotheses}
& The agent fails to consider plausible alternative causes or fault types.
& \\

\midrule

\textbf{Temporal--causal mismatch}
& The inferred causal chain violates event order or expected propagation.
& \multirow{4}{*}{\centering\parbox[c]{0.14\textwidth}{\centering Analysis Structure}} \\

\textbf{Unsupported causal leap}
& The agent asserts causal links not supported by evidence or topology.
& \\

\textbf{Insufficient verification}
& The conclusion is maintained with weak, indirect, or insufficient evidence.
& \\

\textbf{Belief update failure}
& The agent fails to revise its analysis after contradictory evidence appears.
& \\

\midrule

\textbf{Unstable conclusion}
& The final diagnosis contradicts itself or conflicts with prior reasoning.
& \multirow{2}{*}{\centering\parbox[c]{0.14\textwidth}{\centering Decision Report}} \\

\textbf{Non-convergent reporting}
& The agent fails to reach a decisive RCA result and instead repeats or loops.
& \\

\bottomrule
\end{tabular}
\end{table*}

\subsection{RQ1: Does STAR improve end-to-end RCA performance compared with the base RCA agent baselines?}
Table~\ref{tab:main_results_star_models} reports the overall performance of different workflows and foundation models on two datasets, covering both \emph{root cause localization} and \emph{fault type classification}.

\paragraph{Overall comparison.}
Across both Dataset~A and Dataset~B, the proposed STAR-enhanced workflows consistently outperform their corresponding baselines. In particular, \textbf{mABC+STAR} achieves the best overall performance on nearly all localization metrics and most classification metrics, while \textbf{RCAgent+STAR} consistently ranks second, followed by the original \textbf{mABC} and \textbf{RCAgent}. This ordering is stable across different foundation models, demonstrating that the gains mainly come from the stage-attributed correction and replay mechanism rather than from a particular backbone model.

\paragraph{Root cause localization performance}
STAR brings substantial improvements to root cause localization on both datasets. On Dataset~A, with GPT-5 as the backbone, mABC improves from 37.2\%/45.1\%/47.8\% to 56.2\%/66.1\%/70.8\% in Acc@1/3/5 after incorporating STAR, corresponding to absolute gains of +19.0, +21.0, and +23.0 points, respectively. Similarly, RCAgent improves from 23.8\%/30.1\%/32.6\% to 48.7\%/58.4\%/62.5\%, yielding even larger gains of +24.9, +28.3, and +29.9 points. A similar trend is observed on Dataset~B. For example, mABC+STAR with GPT-5 reaches 58.2\%, 68.6\%, and 73.9\% on Acc@1/3/5, compared with 38.7\%, 47.0\%, and 51.1\% for the original mABC. These results indicate that STAR substantially improves the agent's ability to recover the true root cause within both top-1 and top-$k$ predictions.

\paragraph{Fault type classification performance.}
The performance gains are equally evident in fault type classification. On Dataset~A, mABC+STAR (GPT-5) raises MiF1/MaF1 from 0.5103/0.4767 to 0.7204/0.6867, while RCAgent+STAR improves from 0.3755/0.3406 to 0.6555/0.6165. On Dataset~B, the corresponding gains are from 0.5645/0.5338 to 0.7657/0.7333 for mABC, and from 0.4309/0.4006 to 0.6976/0.6663 for RCAgent. These improvements suggest that STAR not only helps identify the faulty service more accurately, but also produces more discriminative intermediate reasoning patterns that benefit downstream fault category recognition.

\paragraph{Effect of STAR on different base workflows.}
An interesting observation is that STAR yields larger relative gains on the weaker baseline, RCAgent, than on mABC. This trend is consistent on both datasets and across all three models. We attribute this to the fact that weaker baselines are more prone to stage-level reasoning errors, making them benefit more from stage attribution, targeted correction, and replay. In contrast, mABC already provides a stronger initial reasoning trajectory, so STAR mainly serves as a refinement mechanism, further improving robustness and top-$k$ localization.

\paragraph{Comparison across foundation models.}
Regarding the foundation models, \textbf{GPT-5} achieves the strongest overall performance in most settings, especially on Acc@1/3/5 and micro-averaged classification metrics. However, \textbf{Qwen3-max} is highly competitive and surpasses GPT-5 on several macro-level metrics, such as MaRe and MaF1 in the STAR-enhanced setting on both datasets. \textbf{Gemini2.5pro} generally performs slightly below the other two models, but still benefits significantly from STAR. This pattern indicates that while stronger reasoning backbones improve the absolute ceiling, the effectiveness of STAR is robust across diverse LLM families.

\paragraph{Cross-dataset observation.}
The same ranking trend holds for both datasets, but absolute performance on Dataset~B is generally higher than on Dataset~A. Since Dataset~B contains real production incidents from a fixed enterprise platform, its fault patterns may exhibit relatively stronger operational regularity than the large-scale public benchmark in Dataset~A. Nevertheless, the consistent improvements across both datasets demonstrate the generality of STAR under both controlled fault injection and real-world incident environments.

\subsection{RQ2: Can STAR precisely identify the decisive faulty stage?}
\begin{figure}[htbp]  
    \centering        
    
    \begin{subfigure}[b]{0.4\textwidth}  
        \centering                        
        \includegraphics[width=\linewidth]{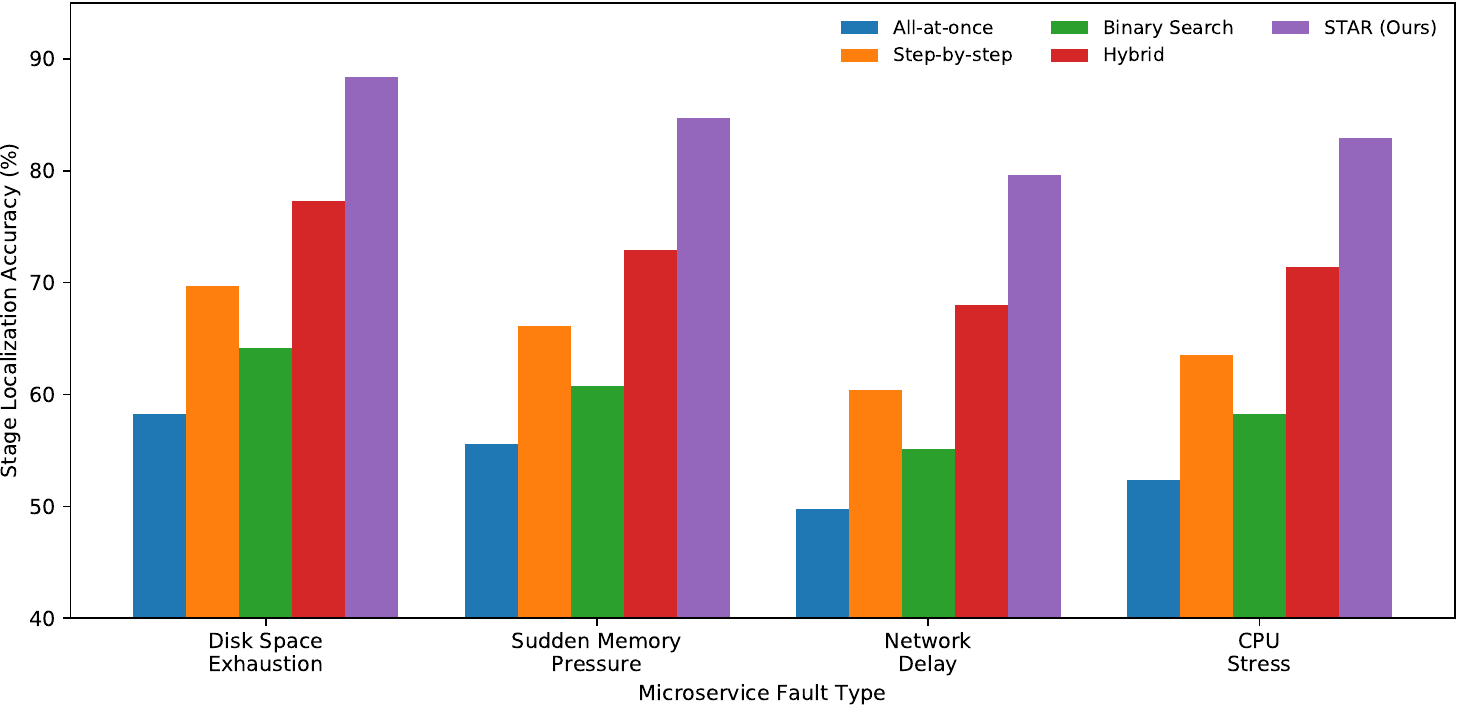}  
        \caption{Comparison Across Microservice Fault Types}
        \label{fig:rq21}               
    \end{subfigure}
    \hfill  
    
    \begin{subfigure}[b]{0.4\textwidth}
        \centering
        \includegraphics[width=\linewidth]{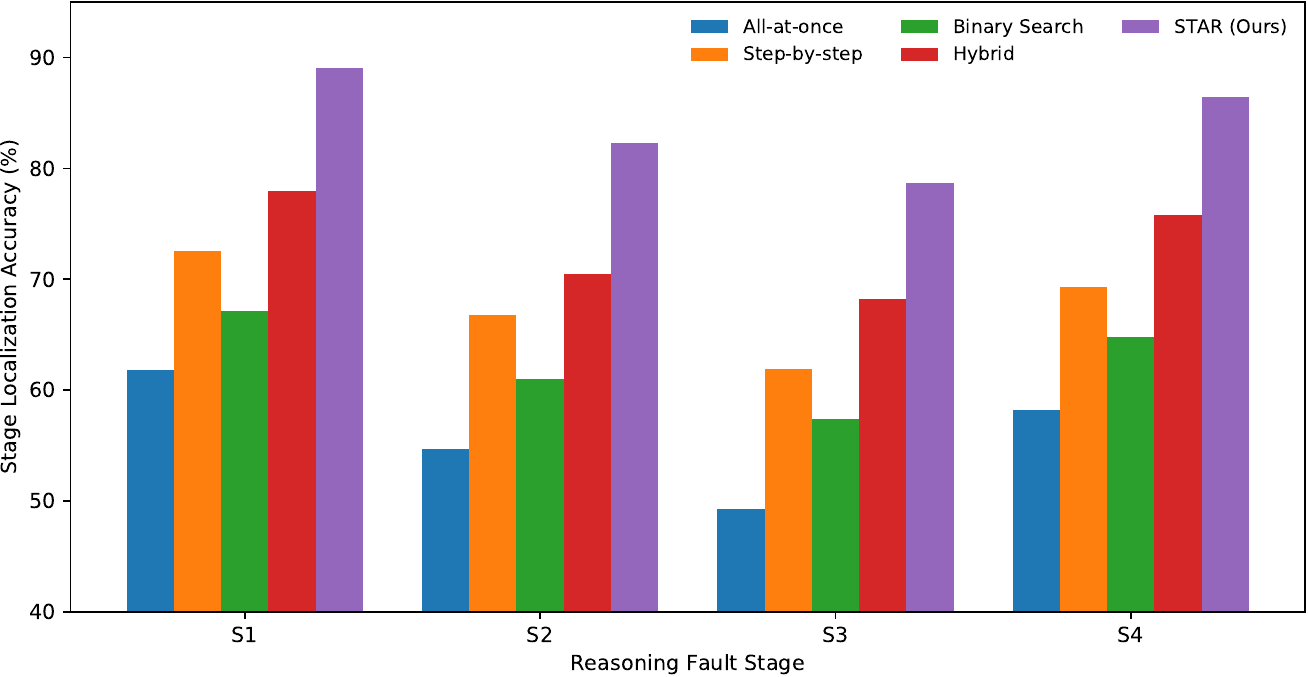}
        \caption{Comparison Across Reasoning Fault Stages}
        \label{fig:rq22}
    \end{subfigure}
    
    \caption{Comparison Across Microservice Fault Stages and Fault Types}
    \label{fig:side_by_side}             
\end{figure}
To answer RQ2, we compare STAR with four adapted failure-attribution baselines inspired by prior work on automated failure attribution for LLM multi-agent systems~\cite{whoandwhen2025}: \textbf{All-at-once}, \textbf{Step-by-step}, \textbf{Binary Search}, and a \textbf{Hybrid} variant.
That work formulates failure attribution as identifying the failure-responsible agent and the decisive error step, and shows that all-at-once is stronger for agent-level attribution, while step-by-step is generally better for precise step-level localization, with binary search lying in between; a hybrid strategy further improves step-level prediction at higher computational cost.
Following this idea, we adapt these methods to our RCA setting by asking each method to directly predict the decisive faulty stage among $S_1$--$S_4$, rather than the exact turn index in a generic multi-agent log.

To answer RQ2, we evaluate decisive-stage localization from two perspectives: \emph{microservice fault type} and \emph{reasoning fault stage}. Specifically, we consider four representative service faults, including \emph{Disk Space Exhaustion}, \emph{Sudden Memory Pressure}, \emph{Network Delay}, and \emph{CPU Stress}, and also inject reasoning faults into each stage $S_1$--$S_4$. The metric is \emph{stage localization accuracy}, i.e., the proportion of cases where the predicted decisive stage matches the ground-truth stage.

As shown in Fig.~\ref{fig:rq21}, STAR consistently outperforms all adapted baselines across all four service fault types, achieving 88.4\%, 84.7\%, 79.6\%, and 82.9\% accuracy, respectively. Among the baselines, Hybrid performs best overall, followed by Step-by-step, Binary Search, and All-at-once. This indicates that STAR benefits from explicitly modeling RCA traces as stage-structured artifacts rather than treating them as generic long-form interaction logs.

Fig.~\ref{fig:rq22} further reports the results across different reasoning fault stages. STAR achieves 89.1\%, 82.3\%, 78.7\%, and 86.4\% on $S_1$--$S_4$, outperforming the strongest baseline by 10+ points on all stages. The same difficulty pattern is observed across methods: $S_1$ and $S_4$ are easier to localize, while $S_2$ and especially $S_3$ are harder, since hypothesis drift and analysis errors are more entangled with intermediate reasoning. Overall, these results show that STAR can reliably identify the decisive faulty stage and that stage-structured attribution is more effective than directly adapting generic failure-attribution methods to RCA.

\subsection{RQ3: How many repair iterations are typically needed before STAR reaches a corrected diagnosis?}
\begin{figure}[t]
    \centering
    \includegraphics[width=0.8\columnwidth]{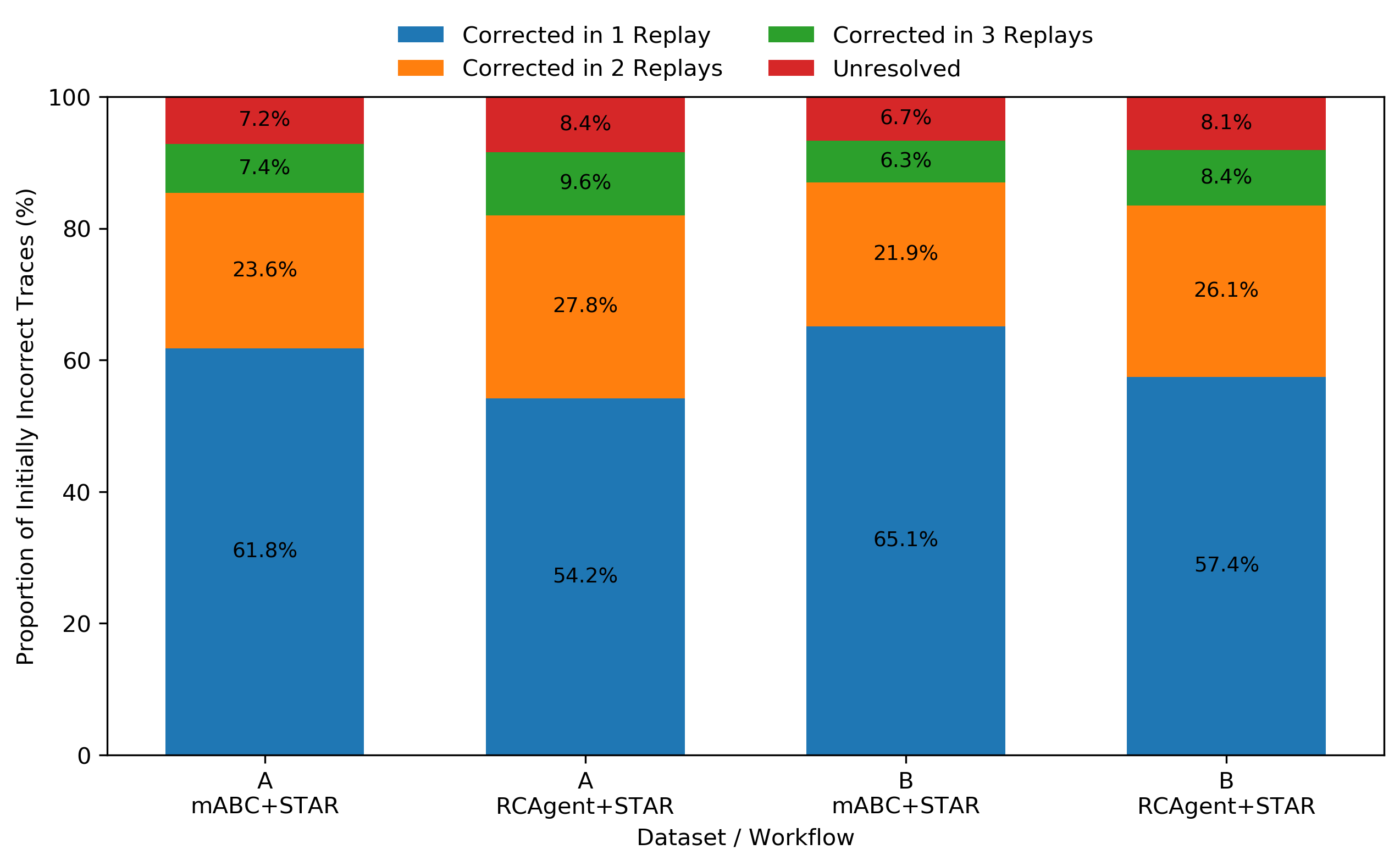}
    \caption{Counterfactual repair iteration distribution when STAR is applied to initially incorrect traces generated by the baseline agents.}
    \label{fig:rq3_repair_iters}
\end{figure}

To answer RQ3, we analyze the repair efficiency of STAR on initially incorrect RCA traces under two workflows, \textbf{mABC+STAR} and \textbf{RCAgent+STAR}, on both Dataset~A and Dataset~B. The statistics in RQ3 are computed on \emph{incorrect traces originally generated by the baseline agents without repair}; the reported replay counts indicate how many STAR repair rounds would be needed \emph{if STAR were applied afterward}. 
For each erroneous trace, STAR iteratively performs decisive-stage localization, stage patching, and downstream replay until the diagnosis is corrected or the replay budget is exhausted. 
We set the maximum number of repair iterations to 3 and report the proportions of cases corrected after the first, second, and third replay, together with the unresolved cases.

Figure~\ref{fig:rq3_repair_iters} shows that most incorrect traces are repaired within the first two replay rounds across all settings. 
On Dataset~A, mABC+STAR corrects 61.8\% of the erroneous traces after a single replay, compared with 54.2\% for RCAgent+STAR. 
After two replays, the cumulative correction ratios further increase substantially, while only 7.2\% and 8.4\% of the cases remain unresolved, respectively. 
A similar trend is observed on Dataset~B, where mABC+STAR fixes 65.1\% of the cases in the first replay and RCAgent+STAR fixes 57.4\%, with unresolved cases remaining below 10\% for both workflows.

Overall, these results indicate that STAR usually requires only a small number of targeted replays to recover a correct diagnosis. 
Moreover, the stronger base workflow (mABC) benefits more from one-shot repair, whereas RCAgent+STAR more often relies on a second replay, suggesting that better initial reasoning trajectories make stage-level patching more immediately effective.

\subsection{RQ4: How much does STAR benefit from its key components, particularly Fast/Slow Routing and Decisive Stage Localization via Counterfactual Candidate Evaluation?}

To answer RQ4, we conduct ablation studies on the two key components of STAR: \emph{Fast/Slow Routing} and \emph{Decisive Stage Localization via Counterfactual Candidate Evaluation}.
To isolate component-level effects from backbone variation, we instantiate both \textbf{mABC+STAR} and \textbf{RCAgent+STAR} with GPT-5 in this subsection.
For \emph{Fast/Slow Routing}, we evaluate both downstream RCA quality and repair efficiency using \textbf{Acc@1}, \textbf{Acc@3}, and the average number of repair iterations on initially incorrect baseline traces.
For \emph{Decisive Stage Localization via Counterfactual Candidate Evaluation}, we measure both the average decisive-stage localization accuracy and the downstream RCA performance (\textbf{Acc@1} and \textbf{Acc@3}).

Table~\ref{tab:ablation_routing} shows that \emph{Fast/Slow Routing} consistently improves both repair efficiency and diagnosis quality.
Removing this module increases the average repair iterations and reduces Acc@1/Acc@3 across all datasets and workflows.
For example, on Dataset~A with mABC+STAR, the average repair iterations decrease from 1.78 to 1.41, while Acc@1/Acc@3 improve from 52.4\%/62.8\% to 56.2\%/66.1\%.
This indicates that Fast/Slow Routing is not merely a computational optimization, but also improves repair quality by matching correction depth to trace difficulty.

Table~\ref{tab:ablation_counterfactual} shows that removing \emph{Decisive Stage Localization via Counterfactual Candidate Evaluation} substantially degrades both stage localization accuracy and downstream RCA performance.
For instance, on Dataset~B with mABC+STAR, stage localization accuracy drops from 88.6\% to 80.3\%, while Acc@1/Acc@3 decrease from 58.2\%/68.6\% to 54.8\%/64.7\%.
Overall, the ablation results confirm that Fast/Slow Routing mainly improves repair efficiency, whereas Counterfactual Candidate Evaluation is crucial for accurate stage attribution and stronger root cause localization.

\begin{table}[t]
\centering
\caption{Ablation on \emph{Fast/Slow Routing}. Lower Avg. Iters is better.}
\label{tab:ablation_routing}
\setlength{\tabcolsep}{2.8pt}
\renewcommand{\arraystretch}{1.05}
\footnotesize
\begin{tabular}{c c c c c}
\toprule
Data & WF & Variant & Avg. Iters$\downarrow$ & Acc@1 / Acc@3 \\
\midrule
\multirow{4}{*}{$A$}
& \multirow{2}{*}{mABC+S}
& w/o F/S  & 1.78 & 52.4 / 62.8 \\
& & Full    & \textbf{1.41} & \textbf{56.2 / 66.1} \\
\cmidrule(lr){2-5}
& \multirow{2}{*}{RCA+S}
& w/o F/S  & 1.86 & 44.6 / 54.3 \\
& & Full    & \textbf{1.51} & \textbf{48.7 / 58.4} \\
\midrule
\multirow{4}{*}{$B$}
& \multirow{2}{*}{mABC+S}
& w/o F/S  & 1.69 & 54.3 / 64.5 \\
& & Full    & \textbf{1.37} & \textbf{58.2 / 68.6} \\
\cmidrule(lr){2-5}
& \multirow{2}{*}{RCA+S}
& w/o F/S  & 1.80 & 46.2 / 56.8 \\
& & Full    & \textbf{1.47} & \textbf{50.1 / 60.7} \\
\bottomrule
\end{tabular}
\end{table}
\begin{table}[t]
\centering
\caption{Ablation on \emph{Counterfactual Candidate Evaluation}.}
\label{tab:ablation_counterfactual}
\setlength{\tabcolsep}{2.6pt}
\renewcommand{\arraystretch}{1.05}
\footnotesize
\begin{tabular}{c c c c c}
\toprule
Data & WF & Variant & Stage Acc. & Acc@1 / Acc@3 \\
\midrule
\multirow{4}{*}{$A$}
& \multirow{2}{*}{mABC+S}
& w/o CCE & 78.4 & 50.7 / 61.9 \\
& & Full   & \textbf{86.9} & \textbf{56.2 / 66.1} \\
\cmidrule(lr){2-5}
& \multirow{2}{*}{RCA+S}
& w/o CCE & 75.1 & 42.3 / 49.6 \\
& & Full   & \textbf{83.7} & \textbf{48.7 / 58.4} \\
\midrule
\multirow{4}{*}{$B$}
& \multirow{2}{*}{mABC+S}
& w/o CCE & 80.3 & 54.8 / 60.2 \\
& & Full   & \textbf{88.6} & \textbf{58.2 / 68.6} \\
\cmidrule(lr){2-5}
& \multirow{2}{*}{RCA+S}
& w/o CCE & 77.6 & 44.1 / 49.9 \\
& & Full   & \textbf{85.9} & \textbf{50.1 / 60.7} \\
\bottomrule
\end{tabular}
\end{table}
\section{Related Works}

\subsection{RCA in Microservices}
Root cause analysis (RCA) in microservices has evolved from correlation- and graph-based methods to multimodal learning over metrics, logs, traces, and topology. Early efforts such as CauseInfer, FacGraph, and MS-Rank explored dependency-aware and correlation-based diagnosis~\cite{causeinfer2014,facgraph2018,msrank2019}, while more recent systems such as MicroHECL, Eadro, Nezha, MRCA, and Trace-based RCA improve localization robustness by jointly modeling observability signals and service dependencies~\cite{microhecl2021,eadro2023,nezha2023,mrca2024,tracemd2024,tracerca2021}. Other recent studies further investigate operation-aware, event-causal, and sparse-observability RCA settings~\cite{micronet2024,tracenet2023,chainofevent2024,sparserca2024}. 

\subsection{LLM-based RCA Agents}
Recent work has begun to cast RCA as an agentic reasoning task using LLMs, where agents collect evidence, generate hypotheses, and produce diagnostic decisions~\cite{mabc2024,flowofaction2025}. More generally, multi-agent LLM frameworks such as ReAct and AutoGen provide a foundation for structured tool use, staged reasoning, and agent collaboration in complex operational tasks~\cite{react2023,autogen2023}. Yet such systems remain vulnerable to reasoning failures such as evidence omission, hypothesis drift, and decision inconsistency, while lacking an explicit mechanism to localize and repair the faulty reasoning stage.

\section{Conclusion}

This paper presented \textbf{STAR}, a stage-attributed correction framework for repairing erroneous reasoning traces in LLM-based RCA agents.
Instead of treating RCA failure as a monolithic end-to-end error, STAR decomposes the diagnostic process into four structured stages and performs \emph{decisive stage localization}, \emph{stage-specific patching}, and \emph{targeted replay}.
Built on top of LangGraph, STAR leverages node-level replay and structured stage artifacts to support controllable and efficient repair.

Experiments on both a public large-scale benchmark and a real-world production dataset demonstrate that STAR consistently improves root cause localization and fault type classification across different workflows and backbone models.
Our results further show that STAR can identify the decisive faulty stage with high precision, repair most initially incorrect traces within one or two replay rounds, and benefit significantly from both Fast/Slow Routing and counterfactual candidate evaluation.
These findings suggest that explicitly modeling \emph{where} an RCA agent fails is an effective path toward reliable and repairable agentic diagnosis.

Future work includes extending stage attribution from four coarse stages to finer-grained sub-stage or tool-level repair, making replay policies more adaptive by jointly optimizing repair quality, latency, and token cost, and generalizing STAR beyond microservice RCA to other structured agent workflows involving evidence gathering, hypothesis formation, analysis, and decision making.

\vspace{12pt}
\color{red}

\end{document}